\pdfoutput=1

\documentclass[11pt]{article}

\usepackage[]{ACL2023}
\usepackage{CJKutf8}
\usepackage[ruled,vlined,linesnumbered]{algorithm2e}
\usepackage{babel,blindtext}
\newcommand{\method}{LLMRefine\xspace}
\usepackage{times}
\usepackage{latexsym}
\usepackage{graphicx}
\usepackage[T1]{fontenc}

\usepackage[utf8]{inputenc}

\usepackage{microtype}

\usepackage{inconsolata}
\usepackage{booktabs}
\usepackage{adjustbox}

\usepackage{amsmath,amsfonts,bm}

\def\eqref#1{equation~\ref{#1}}

\def\1{\bm{1}}

\DeclareMathAlphabet{\mathsfit}{\encodingdefault}{\sfdefault}{m}{sl}
\SetMathAlphabet{\mathsfit}{bold}{\encodingdefault}{\sfdefault}{bx}{n}

\usepackage{multirow}

\title{\method: Pinpointing and Refining Large Language Models via Fine-Grained Actionable Feedback}

\author{Wenda Xu,\thanks{\:\,\, Work done during a Google internship}$^{\:\,\,\dagger}$ Daniel Deutsch,$^\ddag$ Mara Finkelstein,$^\ddag$ Juraj Juraska,$^\ddag$ Biao Zhang,$^\ddag$\\\textbf{Zhongtao Liu,$^\ddag$ William Yang Wang,$^\dagger$ Lei Li,\textsuperscript{\P}\and Markus Freitag$^\ddag$}
\\
$^\dagger$University of California, Santa Barbara,
$^\ddag$Google, \textsuperscript{\P}Carnegie Mellon University\\
\texttt{wendaxu@cs.ucsb.edu}, \texttt{\{dandeutsch,freitag\}@google.com}}

\begin{document}
\maketitle

\begin{abstract}
Recent large language models (LLM) are leveraging human feedback to improve their generation quality. However, human feedback is costly to obtain, especially during inference.
In this work, we propose \textbf{\method}, an inference time optimization method to refine LLM's output. The core idea is to use a learned fine-grained feedback model to pinpoint defects and guide LLM to refine them iteratively. Using original LLM as a proposal of edits, \method searches for defect-less text via simulated annealing, trading off the exploration and exploitation.
We conduct experiments on three text generation tasks, including machine translation, long-form question answering (QA), and topical summarization. \method consistently outperforms all baseline approaches, achieving improvements up to 1.7 MetricX points on translation tasks, 8.1 ROUGE-L on ASQA, 2.2 ROUGE-L on topical summarization.


\end{abstract}

\section{Introduction}
\label{sec:intro}

\begin{figure}[!htb]
    \includegraphics[width=\linewidth]{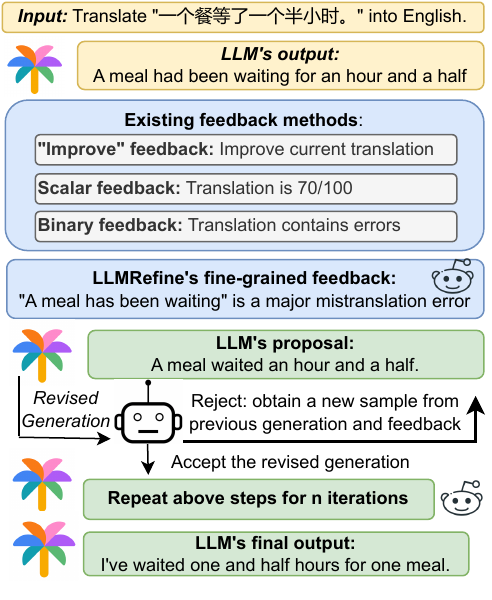}
    \caption{An overview of our \method: We start from LLM's initial generation and iteratively refine the generation, based on fine-grained actionable feedback. We use a simulated annealing technique to accept or reject the proposed revision at each step.}
    \label{fig:pipeline}
\end{figure}





In recent years, large language models (LLMs) have shown impressive performance on various text generation tasks \cite{brown2020language, anil2023palm}.
Critical to their success has been the ability to incorporate human feedback into the learning process \cite{ouyang2022training}.

Nevertheless, human feedback is costly to collect, especially at inference time when the model provides new, unseen input.
In the meanwhile, automatic text generation evaluation metrics for a variety of tasks are rapidly improving \cite{sellam-etal-2020-bleurt,xu2022errors, rei-etal-2020-comet, xu-etal-2023-sescore2,xu2023instructscore}. Can we use one of these metrics to rectify LLM's generation?

In this work, we propose \method, an inference-time optimization method to improve the quality of generated text.
Our \method starts with LLM's initial output, then uses a learned error pinpoint model to provide fine-grained feedback about the location and type of defects in the text. We then use a refinement model (same or another LLM) to follow the feedback instruction and generate candidate text.
The fine-grained feedback provides more much precise information about what exactly is wrong in the generated text, resulting in higher quality revision.



However, due to the large search space, the refinement model is imperfect; it often fails to correct all of the errors identified by the feedback model in one iteration \cite{madaan2023selfrefine}.
We formulate the iterative refinement procedure into a local search problem. It alternates between the feedback generation and refinement in multiple iterations, with the goal of searching for the highest scoring output according to the feedback model. To this end, we develop a simulated annealing technique in \method to trade off between exploring many possible edits and quickly reaching optimal text. Figure~\ref{fig:pipeline} shows overview of our approach.




We evaluate \method on three text generation tasks, including machine translation (WMT \cite{kocmi-etal-2022-findings}), long-form question answering (ASQA \cite{stelmakh-etal-2022-asqa}) and topic summarization \cite{saunders2022selfcritiquing}, because they have a large number of annotated outputs with fine-grained error spans \cite{Freitag_2021, saunders2022selfcritiquing, wu2023fine}.
We use those ratings to train an automatic error pinpoint model that generates a list of error spans along with error categories and severities without the aid of a reference text (which is unavailable during inference) \cite{fernandes2023devil, xu2023instructscore}.
This model serves as our substitute for human feedback. Our experiments show that \method results in higher-quality text compared to baseline methods using other feedback (scalar or binary score) or other search techniques. Our contributions are:

\begin{itemize}
    \item We propose \method, an inference time optimization method to iteratively refine LLM's output with fine-grained actionable feedback, achieving best trade-off between search space and optimal quality.
    \item We demonstrate that \method consistently outperforms all baseline approaches, achieving improvements up to 1.7 MetricX points on translationn tasks, 8.1 ROUGE-L on ASQA and 2.2 ROUGE-L improvements on topical summarization. Humans demonstrate a significant preference for the output of \method over the baseline outputs.  
\end{itemize}

\section{Related Work}
\label{sec:related}
\paragraph{Inference-time Optimization Approach}
We divide techniques for incorporating feedback at inference time into two main techniques \citep{pan2023automatically}: 
generate-then-ranking and feedback-guided generation. The reranking framework involves generating a large set of candidate text outputs from the base model and utilizing a critic model to select the best output. The integration of the critic model can be achieved through chain-of-thoughts \citep{wei2023chainofthought, huang2022large}, binary verifier \citep{li-etal-2023-making}, or a utility function \citep{freitag-etal-2022-high, fernandes-etal-2022-quality}. Our approach is complementary to re-ranking or minimum bayes risk decoding (MBR) strategies, offering additional performance beyond these techniques.

\paragraph{Incorporating Fine-Grained Feedback}
Recent studies have highlighted the benefits of fine-grained error annotation by demonstrating that it can reduce noise in human ratings and increase inter-rater agreement \citep{freitag2021experts} as well as increase automatic metric correlation to human judgments \citet{xu-etal-2022-errors,xu-etal-2023-sescore2,xu2023instructscore}.
One approach to leveraging these benefits is through the use of large language models to self-correct their own output  \citep{madaan2023selfrefine}. Building on this, \citet{chen2023iterative} demonstrate that iterative self-improvement further enhances translation quality. However, despite the unsupervised nature of the self-refine pipeline, the feedback signal is dominated by the large language model's own evaluation capability, which has been shown to be biased towards sentence ordering and its own output \citep{liu2023geval, xu2024perils}. To address this limitation, \citet{wu2023fine} propose a fine-grained reward model that distinguishes rewards at the span-level associating with different error categories.  Orthogonal to this work, we propose an inference time optimization approach to iteratively refine model's output with fine-grained feedback. 



\section{Refinement with Fine-Grained Feedback}
\label{sec:approach}
There are three main components to our framework: a generation model, a feedback model, and a refinement model, each described next.

The generation model produces an initial candidate output $y_i$ given the input $x$.
$x$ and $y_i$ are the source text and a candidate output that is generated by the model.
The feedback model $F$ takes $x$ and $y_i$ and generates some form of feedback $f_i$ that represents the quality of $y_i$, which can be in any form---a scalar value, Boolean, free form natural language, or more.
We assume $f_i$ can always be converted into a scalar quality score via function $s(\cdot)$ (Details of our scoring scheme can be found in Appendix \ref{sec:scoring_scheme}).
Finally, the refinement model uses $x$, $y_i$, and $f_i$ and generates a new, improved output $y_{i+1}$.
As we will discuss in Section~\ref{sec:iterative_refinement}, the loop between the feedback and refinement model can repeat for multiple iterations to further evaluate and update the generated output.

For most of this work, we assume that both the generation and refinement models are an LLM that is 0-shot prompted to perform the respective task (See example prompt in Table \ref{tab:mt_examples}, although we do experiment with different generation models).
The specific prompt for the refinement model depends on the type of feedback being used (See Figure \ref{fig:pipeline}). Since our focus is on the value of fine-grained feedback in the form of an error pinpoint model for text generation, we next describe our feedback model in more detail.

\subsection{An Error Pinpoint Model}

While the majority of text generation evaluation research focuses on predicting a scalar quality score for a text, we instead train an error pinpoint that produces fine-grained feedback on translation quality, similar to InstructScore \cite{xu2023instructscore}.
This is based on the assumption that more specific, actionable feedback will enable the refinement model to generate better output.

The input to our feedback model is the source text $x$ and a hypothesis generation $y_i$.
The feedback model then generates a list of error locations, types, and severities in natural language that are contained in $y_i$.
We model this task as a sequence-to-sequence model and finetune an LLM.
Further implementation details are provided in Section~\ref{sec:setup}.

Training our feedback model requires a set of text with human-annotated error locations, categories, and severities.
For each task that we experiment on, the training data and feedback models are different since the types of errors are task-dependent.
For machine translation, we use MQM annotated data \cite{Mariana2014TheMQ, Freitag_2021}. For long form QA, we use data collected by \citet{wu2023fine}. For topical summarization, we use data collected by \citet{saunders2022selfcritiquing}.

The finegrained feedback model pinpoints the error location and provides detailed error type information and severity level.
This stands in contrast to more traditional evaluation metrics like BLEU, ROUGE or BLEURT that assign scalar scores that represent text generation quality.
Note that because the feedback model operates during inference, our feedback model does not use a reference to evaluate the text. The specific input and output examples for our feedback model can be found in the Table \ref{tab:instructscore_examples}, \ref{tab:instructscore_qa_examples} and \ref{tab:instructscore_summ_examples}.

Once feedback $f_i$ is generated, it is passed to the refinement model via prompting (See Figure \ref{fig:pipeline} for example inputs and outputs to the feedback and refinement model).
Specific implementation and evaluation details of our error pinpoint model are described in Section \ref{sec:err_detection}.

\section{Iterative Refinement as Search}
\label{sec:iterative_refinement}
Although the refinement model receives the output $y_i$ and feedback $f_i$, it is not always guaranteed to generate the best new output in a single step.
Therefore, we experiment with different methods for iterative refinement in which the feedback and refinement loop is repeated until some stopping condition is met.

Iterative refinement can be viewed as a search procedure that is trying to find the optimal $y_i$ for a given $x$, where ``optimal'' is measured by the feedback model.
Specifically, we model iterative refinement as a local search algorithm in which every possible output is a state in the search space, and each step of the search algorithm starts at some state represented by $y_i$ and moves to $y_{i+1}$.
The goal is to find the highest scoring state.

We explore three different local search algorithms, described next.

\subsection{Local Search Algorithms}

Given a current output $y_i$, the local search algorithms begin by sampling a new candidate output $c_i$ from the refinement model given feedback $f_i$.
Then, each algorithm makes a decision about whether it will accept or reject $c_i$ based on some criteria.
If the decision is made to accept $c_i$, then $c_i$ becomes $y_{i+1}$ and the search loop repeats unless the feedback model detects no errors in $y_{i+1}$.
If $c_i$ is rejected, then $y_{i}$ becomes $y_{i+1}$ and the algorithm repeats (i.e., a second candidate is sampled from the refinement model for the same output).
Each of the three following algorithms differs in how it decides whether to accept or reject the candidate output.

\paragraph{Always Accept.}
The ``always accept'' algorithm (AA) will attempt to explore the search space as much as possible by always accepting $c_i$.

\paragraph{Greedy Uphill.}
The greedy uphill (\textsc{Greedy}) algorithm will only accept $c_i$ if the score from the feedback model for $c_i$ is better the score for $y_i$.
In this case, we ensure that the output does not get worse according to the feedback model.

\paragraph{Simulated Annealing.}
The AA and \textsc{Greedy} algorithms each make different trade-offs.
AA will always explore the search space, sometimes at the cost of quality, whereas the \textsc{Greedy} algorithm may do little in terms of search in order to ensure the output quality does not decrease.
Here, we propose a search heuristic based on simulated annealing (SA) \cite{kirkpatrick1983optimization} that tries to combine the strengths of the two approaches.

The SA search algorithm uses a temperature hyperparameter $T$ that controls output diversity and the probability that $c_i$ is accepted.
The probability of acceptance is defined as the following:
\begin{equation}
\label{eq:simulated_annealing}
    p(\textrm{accept} | x, y_i, c_i, T) = \min(1, e^\frac{s(F(c_i)) - s(F(y_i))}{n*T})
\end{equation}
where $n$ is the maximum number of iterations. $i$ is the ith iteration in the pipeline. $T_{i+1} = max(T_i-c*T_i, 0)$. Temperature decays by a constant proportion c. There are two factors contribute to increasing the probability that a candidate is accepted: a high temperature and an improvement in quality according to the feedback model.

At the beginning of the search algorithm, the temperature is set to a high value, allowing the algorithm to explore the search space more liberally.
This allows the SA algorithm to accept a candidate that is potentially worse than the current one, like the AA algorithm.
On each iteration, as the temperature drops, output sample becomes more deterministic and this encourages the model to only accept candidates that are better than the current one, like the \textsc{Greedy} algorithm.
In this way, SA combines the strengths of both alternative search procedures. The pseudocode for the SA algorithm can be found in Algorithm~\ref{alg:search_algo}.

\begin{algorithm}[ht!] \small
  \caption{SA for Iterative Refinement}
  \label{alg:search_algo}
  \KwIn{Input prompt $x$, Feedback model $F$, Base model $M$}

  \textbf{Initialize:} $y_0 \gets greedy\_decode(M(x))$, $T_0$, $n$ \# Initialize candidate, temperature, constant

  \For{$i = 0 .. n$}{
    $f_i \gets F(x, y_i)$ \# generate feedback for the current candidate proposal
    
    $c_{i} \gets Sampling(M(x, y_{i}, f_i))$ \# Sample next candidate based on prior one and feedback 
    
    $p_\textrm{acc} \gets \min(1, e^\frac{s(F(c_i))-s(F(y_i))}{n*T_i})$
    
    \If {\textrm{Accept}} {
        $y_{i+1} \gets c_i$ 
    } 
    \Else {
        $y_{i+1} \gets y_i$
    }
        $T_{i+1} = max(T_i-c*T_i, 0)$ \# update temperature for the next iteration
  }
  \KwOut{Sampled sequence $y_{n}$ with $n$ iterations}
 \end{algorithm}

\section{Experimental Setup}
\label{sec:setup}

Here, we describe the setup for experiments related to implement/evaluate error pinpoint feedback model and implement/evaluate the proposed refinement and iterative refinement procedures.

\subsection{Error Pinpoint Model Implementation and Evaluation}
\label{sec:err_detection}
We leverage the MQM, ASQA and topic summarization datasets to train and meta-evaluate our error pinpoint model.
The model was initialized with PaLM (Bison) LLM and was trained separately for each language pair and each task using WMT'21 MQM data for Zh$-$En (17,185 examples) \cite{freitag-etal-2021-results}, WMT'20 + WMT'21 MQM data for En$-$De (35,340 examples) \cite{freitag-etal-2022-results}, 2853 annotated ASQA examples \cite{wu2023fine} and 17,872 annotated topical summarization examples \cite{saunders2022selfcritiquing}. We use batch size $32$, drop out rate $0.1$, $500$ warm up steps and learning rate $1e-4$ for all languages and tasks. We set maximum prefix length to be $2048$ and maximum decoding step to be $512$. Checkpoint selection was done by selecting the best Pearson correlation on four separate held-out sets, each comprising 500, 500, 500 and 2000 samples respectively. Each set corresponds to Zh-En translation, En-De translation, long form QA and topical summarization. For the reproducing purpose, we perform greedy decoding for the fine-grained feedback generation. We meta-evaluate the error pinpoint model by calculating Pearson correlation and pairwise accuracy between its score and ground-truth human ratings. We evaluate pinpointed error span using character-level precision/recall/F$_1$ scores \citep{blain-EtAl:2023:WMT}. 

We meta-evaluate the error pinpoint model by calculating the correlation between its scores and ground-truth human MQM scores on the WMT'22 English-German and Chinese-English datasets, two benchmark datasets for meta-evaluating metrics.
We calculate a segment score by summing the scores corresponding to the MQM error severity weights that are predicted by our error pinpoint model: 5 for major errors and 1 for minor errors.
We report Pearson and pairwise accuracy with tie calibration \citep{deutsch2023ties} at the segment-level and compare to two state-of-the-art reference-free evaluation metrics, COMETKiwi-QE \cite{rei2022cometkiwi} and BLEURT-QE, a reference-free version of BLEURT \cite{sellam-etal-2020-bleurt} that we trained ourselves on the same data used by COMETKiwi-QE.

To evaluate the actual spans produced by our feedback model, we adopt the character-level precision/recall/F$_1$ that was used by the WMT'23 QE Shared Task.\footnote{\url{https://wmt-qe-task.github.io/}}
The evaluation treats each translation character as a classification problem and micro-averages the results.
We compare our model to AutoMQM \citep{fernandes2023devil} and XCOMET \citep{guerreiro2023xcomet} on the news subset of the WMT'22 MQM dataset because XCOMET uses the remainder of WMT'22 for training (the results on the full dataset are available in Appendix~\ref{appendix:additional_results}.

\subsection{Refinement Implementation and Evaluation}
\paragraph{Datasets \& Evaluation Metrics.}
We evaluate the quality of the text generation produced by our refinement model and search algorithms on the WMT 22 and 23 English-German,  Chinese-English \cite{kocmi-etal-2022-findings}, ASQA \cite{stelmakh-etal-2022-asqa} and topic summarization \cite{saunders2022selfcritiquing} testing set. Each consists of 2037, 557, 1875, 1976, 948 and 284 samples respectively.  The translations are automatically evaluated using MetricX \citep{freitag-etal-2022-results}, a state-of-the-art reference-based evaluation metric for MT. ASQA and summarization are evaluated by ROUGE-L \cite{lin-2004-rouge}.

\paragraph{Baseline Feedback Models.}
We experiment with different baseline forms of feedback for the generated output: 1) \textsc{Improve}: The refinement model is always prompted to improve the translation without being provided any feedback. 2) \textsc{Score-QE}: The refinement model is provided the score from our error pinpoint model and prompted to improve the output. 3) \textsc{Binary-QE}: The refinement model is prompted to improve the output given that if our error pinpoint model spots errors. 4) \textsc{BLEURT-Score-QE}: The refinement model is provided the score from the BLEURT-QE metric and prompted to improve the output. 5) \textsc{BLEURT-Binary-QE}: The refinement model is prompted to improve the output given that the BLEURT-QE score for the output is below a hyperparameter threshold.
The BLEURT feedback models are only available for the MT task.
The prompt used for the refinement model varies based on the type of feedback.
See Appendix Table \ref{tab:feedback_examples} for the specific prompts that were used.

\paragraph{Generation and Refinement Models.}
The majority of our experimentation uses the PaLM-2 (Bison) LLM \citep{anil2023palm} for both initial translation generation as well as the refinement model.
In each case, the LLM was 0-shot prompted for the task.
We also experimented with alternative generation models to understand whether the feedback and refinement models could improve their translations, too.
In particular, we used translations from the systems submitted to the WMT'22 General Machine Translation Task \citep{kocmi-etal-2022-findings}.

\paragraph{Baseline Generation Model.}
We compare the outputs from our proposed refinement models to that of 0-shot prompted PaLM-2.
This model serves as a comparable baseline and allows us to understand how incorporating feedback via the refinement model can improve the original generation quality.

\paragraph{Implementation Details.}
The threshold for determining whether or not an error exists for the \textsc{BLEURT-Binary-QE} model was set to be 0.95 and 1 for Zh-En and En-De, respectively.\footnote{
    The BLEURT-QE scores are mostly between 0 and 1, but high-quality translations often receive a score $>1$.
}
They were chosen using the held-out WMT'21 test set \cite{akhbardeh-etal-2021-findings}.

For the reproducing purpose, we conduct all single step refinement using greedy decoding. For iterative refinement, we use top-k sampling, with $k$=$40$. For the uphill and always accept algorithms, temperature was set to default value $0.8$. For the SA search algorithm, the initial temperature was set to $0.8$ and is reduced by 10\% on each iteration. We experiment different normalization constants from $1$ to $10$ on our development set WMT21 and choose the best performed constant $4$ during simulated annealing. For iterative improvement, we set the maximum number of iterations $n$ to be 10.

\section{Results}
\label{sec:exps}

We explore several research questions in our experiments: 1) How well does our error pinpoint model align with human annotations of translation quality? 2) Does fine-grained feedback result in better downstream translations than more coarse feedback? 3) Can the feedback and refinement models be used to improve translations generated by unrelated text generation models? 4) Does the iterative refinement improve the generated translation quality?

\begin{table}[t]\centering
\resizebox{0.48\textwidth}{!}{
\scriptsize
\begin{tabular}{lcccc}
\toprule
\multicolumn{1}{c}{\bf Metric} &\multicolumn{2}{c}{\bf Zh-En} &\multicolumn{2}{c}{\bf En-De} \\
& \bf Acc. & \bf $r$ & \bf Acc. \bf &$r$ \\
\midrule
COMETKiwi-QE & 0.516 & 0.509 & 0.583 & \bf 0.432 \\
BLEURT-QE & 0.523 & 0.385 & 0.591 & 0.392 \\
Error Pinpoint Model & \bf 0.535 & \bf 0.516 & \bf 0.601 & 0.394 \\

\bottomrule
\end{tabular}
}
\caption{Segment-level accuracy (after performing tie calibration) and Pearson's $r$ of our error span detection model compared to other reference-free evaluation metrics on the WMT'22 zh-en and en-de datasets.}
\label{tab:feeback_results}
\end{table}
\begin{table}[t]
    \centering
    \begin{adjustbox}{width=0.95\columnwidth}
    \begin{tabular}{lcccccc}
        \toprule
        \multirow{2}{*}{\bf Metric} & \multicolumn{3}{c}{\bf En-De} & \multicolumn{3}{c}{\bf Zh-En} \\
        \cmidrule{2-7}
        & \bf P & \bf R & \bf F$_1$ & \bf P & \bf R & \bf F$_1$ \\
        \midrule
        AutoMQM (Bison) & 0.05 & 0.58 & 0.09 & 0.10 & 0.17 & 0.13 \\
        XCOMET-XXL & 0.24 & 0.38 & 0.29 & 0.15 & 0.57 & 0.24 \\
        Error Pinpoint Model & \textbf{0.28} & 0.21 & 0.24 & \textbf{0.30} & 0.31 & \textbf{0.30} \\
        \bottomrule
    \end{tabular}
    \end{adjustbox}
    \caption{Character-level precision/recall/F$_1$ of different error span tagging models (XCOMET-XXL is a reference-based metric and AutoMQM and error pinpoint are reference-free metrics).
    Our error pinpoint model has the highest precision compared to others, even with reference-based XCOMET. This implies that our predicted error spans are most reliable.}
    \label{tab:charf1_news}
\end{table}
\begin{table*}[!htp]\centering
\resizebox{0.95\textwidth}{!}{
\scriptsize
\begin{tabular}{lrrrrrrr}\toprule
\multicolumn{1}{c}{} &\multicolumn{1}{c}{\textbf{MT22 Zh$-$En}} &\multicolumn{1}{c}{\textbf{MT23 Zh$-$En}} &\multicolumn{1}{c}{\textbf{MT22 En$-$De}} &\multicolumn{1}{c}{\textbf{MT23 En$-$De}} & \multicolumn{1}{c}{\textbf{ASQA}} & \multicolumn{1}{c}{\textbf{Topical Summ}}\\
\midrule
\textit{Baseline}&Metric-X & Metric-X & Metric-X & Metric-X & ROUGE-L & ROUGE-L\\
\textsc{PaLM-2 0-shot} &75.3 &73.8 & 83.1 & 78.3 & 17.6 & 28.7\\
\midrule
\multicolumn{7}{l}{\emph{Feedback Models}} \\
\textsc{Improve} &75.6 & 74.0 & 78.9 & 77.8 & 19.2 & 28.8\\
\textsc{BLEURT-Score-QE} &75.6 & 74.1 & 80.0 & 77.8 & - & - \\
\textsc{BLEURT-Binary-QE} & \textbf{75.9} & 74.1 &  82.3  & 78.9 & - & - \\
\textsc{Score-QE} & 75.6 & 74.0 & 83.2 & 79.0 & 21.9 & 29.4\\
\textsc{BINARY-QE} &75.7 & 74.0 & 83.3 & 79.1 & 21.6 & 29.1\\
\textsc{\method} & \bf 75.9 &\bf 74.2 & \bf 83.5 & \bf 79.3 & \bf 26.1 & \bf 30.5\\
\bottomrule
\end{tabular}
}
\caption{\label{tab:main_results} We include three baseline models using coarse feedback: \textsc{Improve}, \textsc{Binary-QE}, \textsc{Score-QE}, \textsc{BLEURT-Binary-QE}, \textsc{BLEURT-Score-QE} and \method, which is guided by our fine-grained error pinpoint model. All results are obtained through greedy decoding. In Appendix Table \ref{tab:mistral_moe_metricx} and \ref{tab:palm2_comet22}, we report additional results on open sourced LLMs and results of COMET scores to demonstrate the effectiveness of our method on open sourced models.}
\end{table*}


\subsection{Meta-Evaluating the Pinpoint Model}
\label{sec:eval_feedback}
Table~\ref{tab:feeback_results} contains the segment-level meta-evaluation results for our error pinpoint model, BLEURT-QE, and COMET-QE.
In all but one evaluation setting, our feedback model has the best results compared to the strong baseline metrics.
Therefore, we conclude that the feedback model is a state-of-the-art evaluation metric and is a high-enough quality to be used in the rest of our experiments. 

Table~\ref{tab:charf1_news} contains the automatic evaluation of the predicted spans.
Among the metrics, our feedback model achieves the highest Character-level precision on both language pairs and the best Character-level F$_1$ on Chinese-English, making it a suitable candidate for identifying errors that should be corrected during the refinement step of our pipeline. 

We meta-evaluate our error pinpoint model by comparing the gap in downstream translation quality when human-annotated error spans are used. This is to measure the effectiveness of our feedback model in guiding the refinement. We can compare the performance improvements achieved in this way (i.e., with a professional annotator's guidance) to those achieved with our feedback model's guidance (see Human vs.~Inst-QE). What we find is that the performance of refinement with the feedback model is competitive, achieving an average improvement of $2.2$ MetricX in En-De and $2.8$ MetricX in Zh-En, with the scores on average a mere $0.2$ and $0.3$ behind those achieved with oracle human feedback for En-De and Zh-En, respectively. This discovery validates the effectiveness of our automatic feedback in improving the quality of the base translation. You can find input output examples of error pinpoint model for each task at Appendix Table \ref{tab:instructscore_examples},\ref{tab:instructscore_qa_examples} and \ref{tab:instructscore_summ_examples}

\subsection{Fine- vs. Coarse-grained Feedback}
Table~\ref{tab:main_results} compares the quality of the refined translations when different forms of a feedback are used plus the PaLM-2 0-shot baseline quality.

Inadequate feedback could deteriorate the generation.
While always prompting the refinement model to improve (\textsc{Improve}) exhibits better translation performance for WMT'22 and WMT'23 when focusing on Zh$-$En, it results in a significant decline in translation quality for En-De. This highlights the instability of the direct prompting approach.
Similar patterns are observed when examining using only scalar feedback scores from \textsc{BLEURT-Score-QE} and only refining translations when the metric predicts there is an error (\textsc{BLEURT-Binary-QE}). We observe steady performance improvements by adding more detailed feedback at translation, long form question answering and Topical summarization. We argue that the lack of detailed error analysis increases the task difficulty and can't fully elicit LLM's refinement ability.

\begin{table}
    \centering
    \begin{adjustbox}{width=0.98\columnwidth}
    \begin{tabular}{lcccccc}
        \toprule
        \multirow{2}{*}{\bf Model} & \multicolumn{2}{c}{\bf WMT'22} & \multicolumn{2}{c}{\bf WMT'23} & \multicolumn{1}{c}{\bf ASQA} &  
        \multicolumn{1}{c}{\bf Summ}\\
        & \bf Zh-En & \bf En-De & \bf Zh-En & \bf En-De & \bf QA & \bf Summ\\
        \midrule
        PaLM-2 0-shot & 66.1 & 77.0 & 65.7 & 75.1 & 17.6 & 25.2\\
        \textsc{Improve} & 67.7 & 77.1 & 67.5 & 75.9 & 19.2 & 25.5\\
        \textsc{Score-QE} & 67.5 & 77.2 & 67.2 & 76.3 & 21.9 & 26.4\\
        \textsc{Binary-QE} & 67.6 & 77.9 & 67.3 & 76.5 & 21.6 & 26.0\\
        \textsc{\method} & \bf 68.8 & \bf 78.6 & \bf 68.2 & \bf 76.9 & \bf 26.1 & \bf 28.1\\
        \bottomrule
    \end{tabular}
    \end{adjustbox}
    \caption{Fine-grained feedback vs coarse feedback on the examples that are marked as "errors" by our error pinpoint model. MetricX is used for all translation results and ROUGE-L is used for ASQA and summ.}
    \label{tab:subset_improvements}
\end{table}



By contrast, fine-grained feedback from our error pinpoint model delivers significant and consistent improvements:
Using our fine-grained feedback model with a single iteration consistently enhances the quality of the base translation in both Zh$-$En and En$-$De across all four testing sets and achieves the highest performance at ASQA and topic summarization. 


\paragraph{Examining Generations with Errors}
Many of the original outputs from our PaLM-2 generation model are already error free according to our error pinpoint model.
In such cases, no refinement is done and the result does not change, so the magnitude of the MetricX or ROUGE-L improvement made by the refinement model is not well represented. Therefore, we additionally report results on the subset of the WMT'22, '23, ASQA and topical summarization datasets for which our feedback model detected an error
\footnote{This consists of 407/1875, 329/1976, 465/2037, 334/557, 937/948 and 166/284 on the WMT'22 Zh-En, WMT'23 Zh-En, WMT'22 En-De, WMT'23 En-De, ASQA and topical summarization respectively}.

From Table~\ref{tab:subset_improvements}, we observe the improvements are much larger than on the entire dataset as a whole.
For example, on WMT'22 zh-en, the improvement using our error pinpoint model is 2.7 MetricX points compared to 0.6 on the full dataset.
This further demonstrates the effectiveness of our method.
When the feedback model detects an error, the refinement model can make significant quality improvements.



\begin{figure}
    \minipage{0.24\textwidth}
      \includegraphics[width=\linewidth]{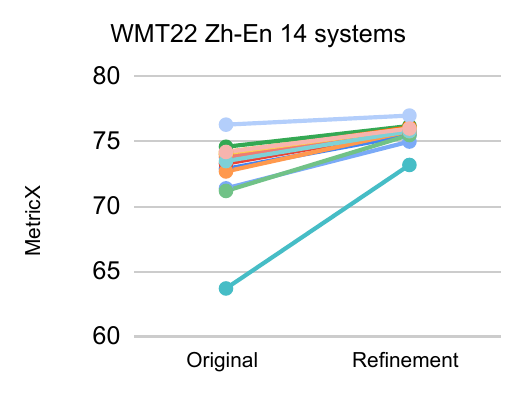}
    \endminipage\hfill
    \minipage{0.24\textwidth}
      \includegraphics[width=\linewidth]{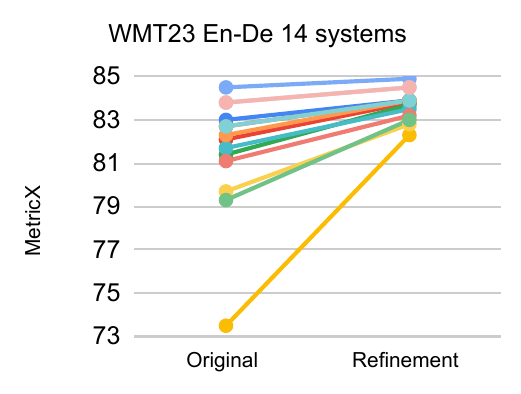}
    \endminipage\hfill
    \caption{MetricX score improvement after one-step refinement of external systems' translations, evaluated on WMT22 Zh-En and En-De.}
    \label{fig:wmt22_system_human_improvements}
\end{figure}

\subsection{Improving Other Source of Generation}

We study the possibility of improving initial translations that come from systems other than PaLM-2, or even improving human translations. We conduct experiments on Zh-En and En-De for 14 submission systems and one set of human translations from WMT22. We performed one step refinement based on fine-grained feedback. 

In Figure~\ref{fig:wmt22_system_human_improvements}, our refinement pipeline consistently improves all of the WMT22 systems, with an average improvement of $2.2$~MetricX in En$-$De and $2.8$ MetricX in Zh$-$En. Notably, it is effective in improving the translation quality of systems that already demonstrated better performance than the PaLM 2 zero-shot translation.

Although the human translations are high-quality, they still contain errors as marked by MQM raters \cite{freitag-etal-2022-results}, therefore, there is room for improvement.
Indeed, we find that our single-step refinement manages to improve even those by as much as $0.8$ MetricX in the Zh-En task, and $0.7$ MetricX in En-De.

\begin{figure}[!htb]
\minipage{0.46\textwidth}
  \includegraphics[width=\linewidth]{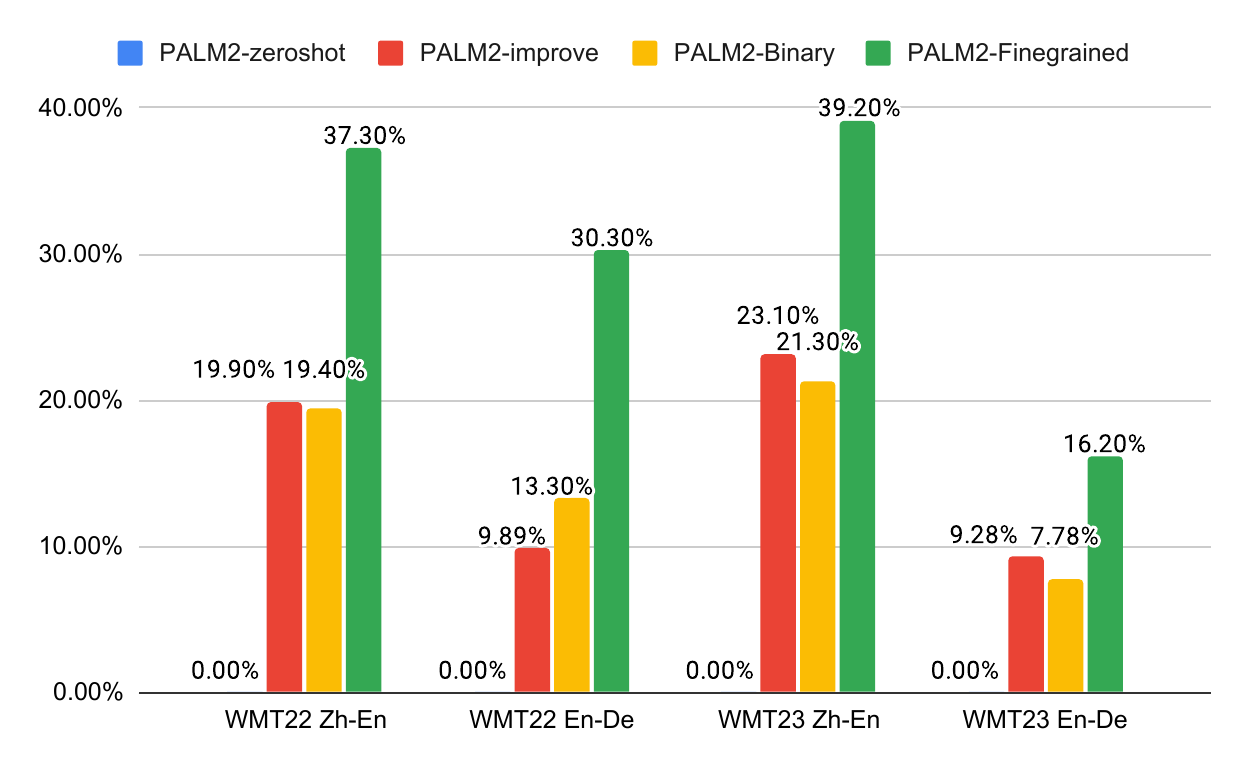}
  \caption{Fine-grained feedback improves the percentage of the corrections.}
  \label{fig:err_inst_qe_percents}
\endminipage\hfill
\minipage{0.49\textwidth}
  \includegraphics[width=\linewidth]{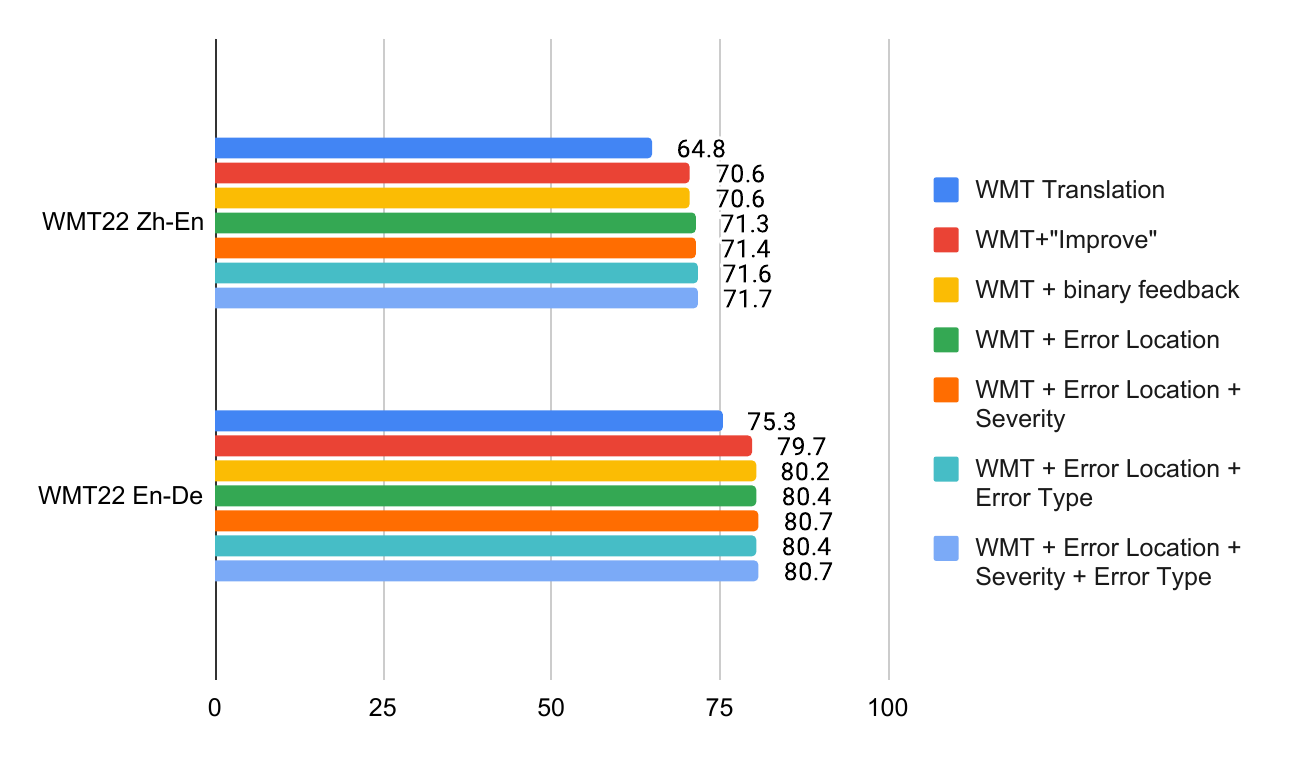}
  \caption{Fine-grained feedback improves the refinement performance}
  \label{fig:ablation_granularity}
\endminipage\hfill
\end{figure}

To further analyze the granularity of fine-grained feedback, we conducted an ablation study on 10647 system outputs for WMT22 Zh$-$En and 6441 system outputs for WMT22 En$-$De, all of which were flagged as containing errors by our feedback model. Specifically, we examined the additive effects of each component (error location, severity, error type) and their contribution to the overall performance. 
Figure~\ref{fig:err_inst_qe_percents} shows that fine-grained feedback significantly improves error correction rate compared to coarse feedback, with a 17\% increase for Zh-En and 13\% for En-De translations, as measured by our error pinpoint model. In Figure~\ref{fig:ablation_granularity}, we observed that providing prompt with error location information significantly improved performance for WMT22 Zh$-$En, resulting in a MetricX improvement of 0.7 compared to only mentioning sentences containing errors. Additionally, severity labels and error types each had their own unique additive effects on the final performance. Finally, when all fine-grained feedback, including error type, location, and severity label, were combined, the joint feedback approach achieved the highest improvements.

\subsection{Iterative Refinement}
\begin{figure}[!htb]
\minipage{0.24\textwidth}
  \includegraphics[width=0.95\linewidth]{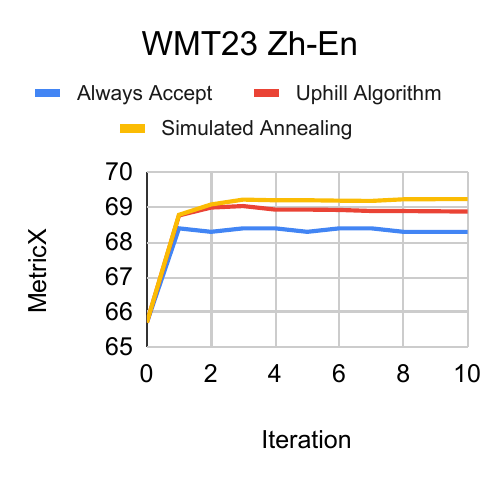}
\endminipage\hfill
\minipage{0.24\textwidth}
  \includegraphics[width=0.95\linewidth]{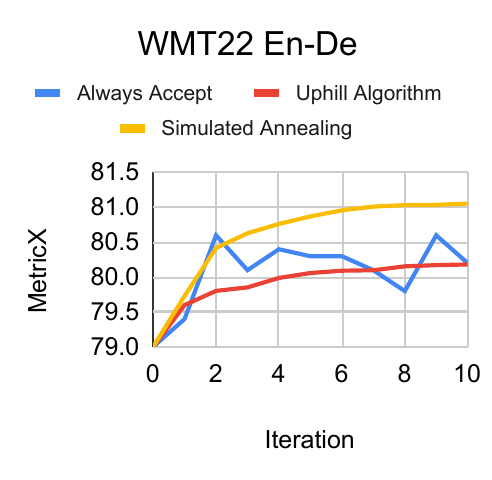}
\endminipage\hfill
\minipage{0.24\textwidth}
  \includegraphics[width=0.95\linewidth]{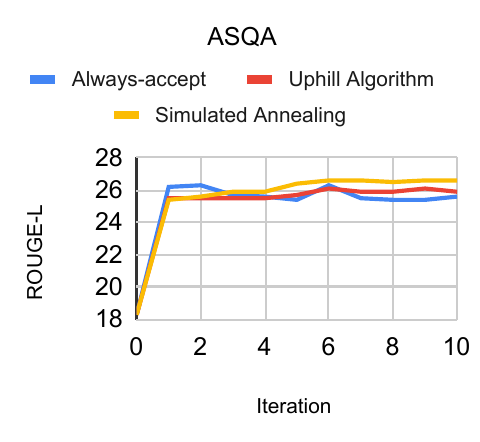}
\endminipage\hfill
\minipage{0.24\textwidth}
  \includegraphics[width=0.95\linewidth]{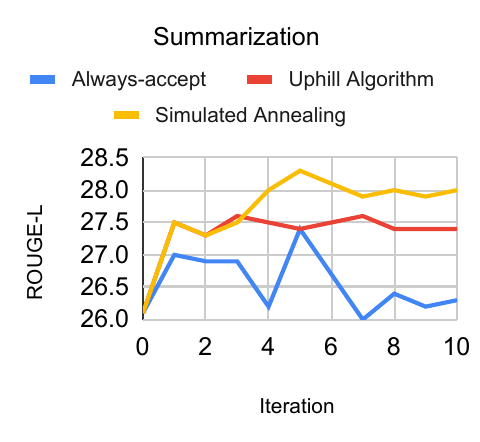}
\endminipage\hfill
\caption{We conducted iterative experiments on WMT23 Zh-En and WMT22 for En-De, ASQA and topical summarization with always accept, greedy uphill and simulated annealing algorithms and report MetricX and ROUGE-L score.}
\label{fig:iterative_study}
\end{figure}

Figure~\ref{fig:iterative_study} contains the results from running the various proposed refinement algorithms for up to 10 iterations. We demonstrate that Always Accept and Greedy Uphill each outperforms another in different test sets due to a trade-off between search space and error feedback. Overall, Simulated Annealing performs best in multi-step refinements.
The figure shows that all three algorithms can result in further performance improvements on top of the initial output.
Notably, we observe that always accepting the output (AA) demonstrates rapid convergence to the maximum, typically requiring only around 1 or 2 iterations. However, it can demonstrate instability of performance (See the fluctuation of the performance in the right figure) as they are lack of a quality selection process.
In contrast, the uphill and simulated annealing techniques yield additional performance improvements over more iterations. We also include detailed iterative results with all tasks for first and fifth iteration at Appendix Table \ref{tab:ier_refine_table}.

\begin{figure}[!htb]
\minipage{0.24\textwidth}
  \includegraphics[width=0.95\linewidth]{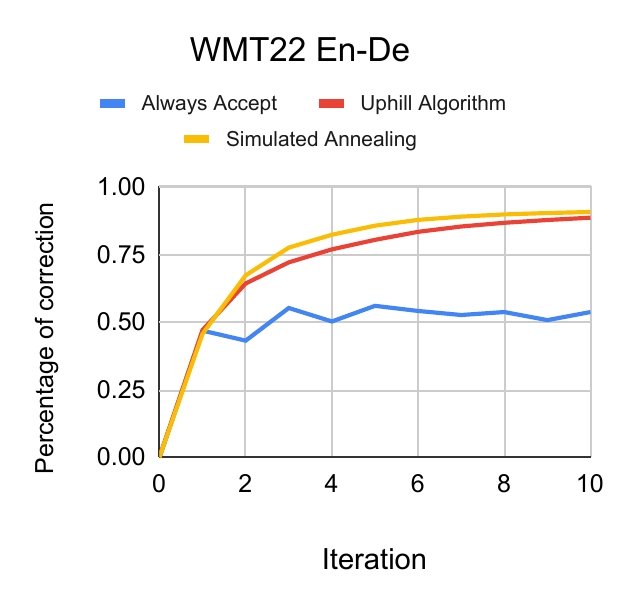}
\endminipage\hfill
\minipage{0.24\textwidth}
  \includegraphics[width=0.95\linewidth]{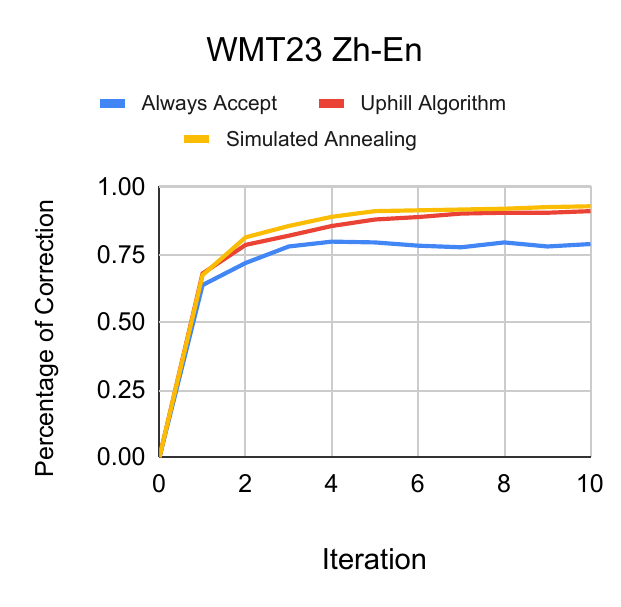}
\endminipage\hfill
\caption{We conducted iterative experiments on WMT22 En-De and WMT23 Zh-En with Always Accept, Uphill and Simulated annealing algorithms and report correction rate of error pinpoint.}
\label{fig:compare_search}
\end{figure}

\begin{figure}[!htb]
\minipage{0.24\textwidth}
  \includegraphics[width=0.95\linewidth]{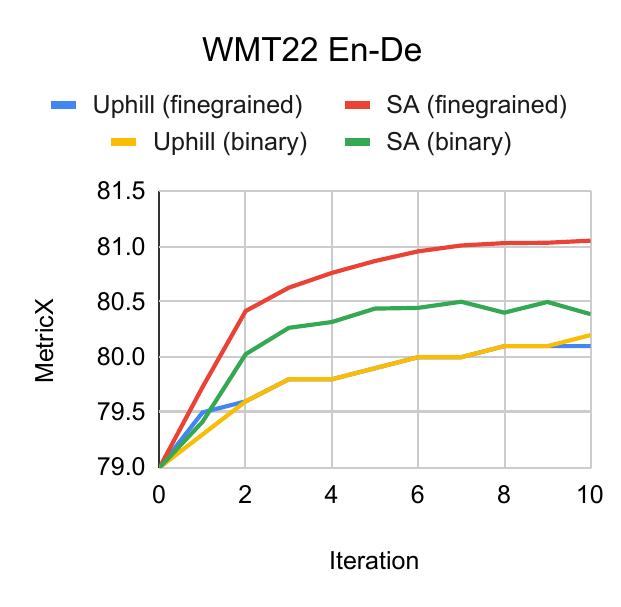}
\endminipage\hfill
\minipage{0.24\textwidth}
  \includegraphics[width=0.95\linewidth]{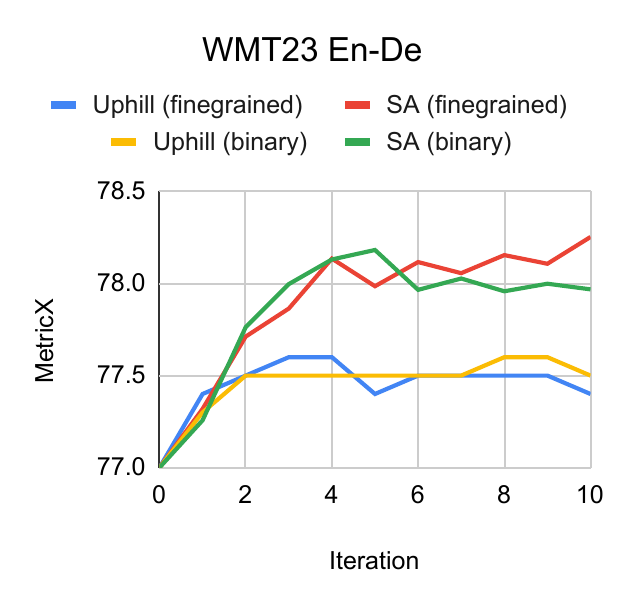}
\endminipage\hfill
\caption{We study whether finegrained feedback can find better candidate generation compared to binary feedback. We conducted experiment on WMT22 and WMT23 at En-De with uphill and simulated annealing and report MetricX.}
\label{fig:annealing_better_proposal}
\end{figure}

\paragraph{Simulated Annealing finds better candidates.}
In Figure \ref{fig:iterative_study} and \ref{fig:compare_search}, we present a comparison of three search algorithms on WMT22 En$-$De and WMT23 Zh$-$En, focusing on their performance in terms of MetricX/ROUGE-L and correction rate improvements (identified by our error pinpoint model). Our observations indicate that during the first iteration, simulated annealing may identify samples that achieve lower MetricX scores and corrects a smaller number of examples compared to uphill algorithm. However, owing to the diverse candidate proposals generated by SA under the measure of performance improvements, by the second to forth iteration, it already identifies samples that achieve higher MetricX scores. The performance gap between the three algorithms widens over the subsequent four or five iterations, ultimately resulting in a superior candidate proposal at the end of the iterations. We provide a concrete case study example in Appendix Table \ref{tab:case_study}. In contrast to always accept, although with full search space, it could occasionally optimize MetricX on WMT22 En$-$De or ASQA. It can not fully optimize error correction rate without a quality selection process. On average, simulated annealing improves the uphill algorithm by $0.5$ MetricX on Zh$-$En, $0.9$ on En$-$De and by $0.7$ ROUGE-L on ASQA and $0.9$ ROUGE-L on topical summarization.

In Figure \ref{fig:annealing_better_proposal}, we empirically show that simulated annealing can boost the performance for different types of feedback (binary and fine-grained). Specifically, we show that simulated annealing with both feedback can significantly improve the proposal quality for their counterparts under uphill algorithm. Furthermore, we demonstrate that simulated annealing with fine-grained feedback can achieve the best MetricX score in additional iterations.


\begin{table}
    \centering
    \begin{adjustbox}{width=0.98\columnwidth}
    \begin{tabular}{lcccc}
        \toprule
        \method vs & \bf Win & \bf Neutral & \bf Lose & Win lose ratio\\
        \midrule
        \textsc{0-shot} & 38\% & 46\% & 16\% & 2.34\\
        \textsc{Improve} & 39\% & 45\% & 16\% & 2.44\\ 
        \textsc{BLEURT-SCORE-QE} & 41\% & 44\% & 15\% & 2.79\\
        \textsc{BLEURT-binary-QE} & 33\% & 48\% & 19\% & 1.76\\
        \textsc{score-qe} & 33\% & 40\% & 27\% & 1.23\\
        \textsc{binary-qe} & 34\% & 48\% & 18\% & 1.84\\
        \bottomrule
    \end{tabular}
    \end{adjustbox}
    \caption{We conduct head-to-head human evaluation on \method against all baselines for single step refinement at WMT22 En-De. We report percentages of win, neutral and lose and win/lose ratio in the table.}
    \label{tab:human_eval}
\end{table}


\begin{table}
    \centering
    \begin{adjustbox}{width=0.98\columnwidth}
    \begin{tabular}{lcccc}
        \toprule
        Simulated Annealing vs & \bf Win & \bf Neutral & \bf Lose & Win lose ratio\\
        \midrule
        \textsc{Always-Accept} & 38\% & 38\% & 24\% & 1.56\\
        \textsc{Greedy Uphill} & 31\% & 47\% & 22\% & 1.38\\ 
        
        \bottomrule
    \end{tabular}
    \end{adjustbox}
    \caption{We conduct head-to-head human evaluation on our simulated annealing based algorithm against greedy at WMT22 En-De. We report percentages of win, neutral and lose and win/lose ratio in the table.}
    \label{tab:human_eval_sa}
\end{table}


\subsection{Human Evaluation Results}
We conduct head-to-head human evaluation on WMT22 En-De with the samples that marked as errors by our feedback model (465/2037). Professional bilingual annotators were shown the source sentence and asked to rate the quality of test translation against base translation with either win, lose or neutral options. We use the win/lose ratio of test translation against base translation as the final metric. If win/lose ratio is greater than 1, then test translation outperforms base translation. In Table \ref{tab:human_eval}, we compared single step fine-grained refinement against all coarse feedback baselines. We found that fine-grained feedback outperforms all other source of feedback, as win/lose are all above 1. Moreover, we compare simulated annealing (SA) baseline against always-accept (AA) and greedy baseline at 5th iteration. In Table \ref{tab:human_eval_sa}, we find win/lose ratios are 1.56 and 1.38 respectively, indicating SA has superior performance against AA and greedy.

\section{Conclusion}
\label{sec:conclusion}
In this work, we proposed \method for incorporating a fine-grained error pinpoint into a text generation pipeline via refinement-feedback model loop.
We empirically demonstrated that our fine-grained error pinpoint model with simulated annealing achieves superior performance compared to baseline feedback models and other search algorithms across three text generation tasks. Lastly, our experimental findings are further solidified by the human evaluation study. Humans demonstrate a significant preference for the output of \method over the baseline outputs.  

\newpage

\section{Limitations}
In this work, we demonstrate that our error pinpoint can achieve comparable Pearson correlation and segment-level accuracy to source based SOTA metrics in Machine translation at WMT22 En-De and Zh-En. Despite the great performance, we also demonstrate the trade-off between precision and recall at Table \ref{tab:charf1_news}. Although achieving higher precision score compared to baseline metrics, our error pinpoint has lower recall. Future work can consider a better pipeline to boost the recall of our error pinpoint while maintaining a reasonably high precision score. Ultimately, this may significantly improve the quality of corrections after iterative refinement.

While \method in theory can be applied to landscape of instruction-fine-tuned large language models, it would be noteworthy to mention that the efficiency may differ when applied with models that lack instruction following capability. Training a large language model with instruction, feedback following ability can be great a future direction to mitigate this issue. 

\section{Ethical Statement}
All the training data of our error pinpoint model is publicly available. We ascertain that the feedback data that is annotated by  human labors do not contain risk or toxic content. We used an internal, proprietary tool to collect human evaluation data. The annotators were compensated fairly and did not have to disclose any personal information during the annotation process. All of the test sets used
in this study are publicly available, and annotators
were allowed to label sensitive information if necessary. The annotators are fully aware that the data which is collected from them will be used for research purposes. Each pair of translations is annotated by one rater. There are six annotators participated for each test vs base system comparisons. 

\section{Acknowledgement}
This work was supported by the National Science
Foundation award \#2048122. The views expressed
are those of the author and do not reflect the official
policy or position of the US government.



\bibliography{custom}
\bibliographystyle{acl_natbib}

\newpage

\appendix


\section{Feedback Scoring Scheme}
\label{sec:scoring_scheme}
We adopted the same setting as human evaluation respect to each task. For machine translation and long form question answering, we adopt MQM human scoring scheme \cite{Freitag_2021} by counting the total number of errors and assigning the weight penalties to the individual error. We use the same severity level (major and minor) and weight penalty as MQM, assigning major error as penalty $5$ and minor error as penalty $1$. For the sentence without errors, we assign a score of $0$. Similar to MQM scheme \cite{Freitag_2021}, for sentence contains more than five major errors, they all assigned to $-25$. The score range is between $0$ to $-25$. For topical summarization, we adopt the same severity levels as \citet{saunders2022selfcritiquing}'s annotations. In particular, they contain four levels of severity: critical, major, medium and minor. Each response will only be assigned with one error type. Therefore, we score each sentence with $1$ to $5$ rating, with $5$ corresponds to no-error and $1$ corresponds to one critical error. We use those scores and human ratings to compute Pearson correlation and pairwise accuracy for meta-evaluating the feedback model. To ensure fair comparisons at our \textsc{Score-QE} baseline, we normalize different score ranges into $0$ to $100$. We include scoring examples of error pinpoint model at Table \ref{tab:instructscore_examples}, \ref{tab:instructscore_qa_examples} and \ref{tab:instructscore_summ_examples}.

\begin{table*}[!ht]
\centering

\footnotesize
\begin{tabular}{p{15cm}}
\toprule
\textbf{Source:}
\begin{CJK*}{UTF8}{gbsn}
一个餐等了一个半小时。
\end{CJK*}\\
\textbf{Reference:} I’ve waited one and a half hours for one meal.\\
\textbf{Output:} A meal had been waiting for an hour and a half.\\
\midrule

\textbf{1st candidate generation: } A meal had been waiting for an hour and a half.\\
\textbf{Feedback: } Error type: mistranslation Major/minor: major Error location: A meal had been waiting. \\
\textbf{Our Error Pinpoint Model: } -5\\
\textbf{BLEURT: } 0.738\\
\midrule

\textbf{2nd candidate generation: } "A meal had been waiting for an hour and a half." is a major mistranslation error.The correct translation should be: "The meal had been waiting an hour and a half." \\
\textbf{Feedback: } Error type: mistranslation Major/minor: major Error location: A meal had been waiting for an hour and a half.\\
\textbf{Our Error Pinpoint Model: } -5\\
\textbf{BLEURT: } 0.692\\
\midrule

\textbf{3rd candidate generation: } The meal had been waiting for an hour and a half\\
\textbf{Feedback: } Error type: mistranslation Major/minor: major Error location: The meal had been waiting\\
\textbf{Our Error Pinpoint Model: } -5\\
\textbf{BLEURT: }0.781\\
\midrule

\textbf{4th candidate generation: }
A meal waited an hour and a half.\\
\textbf{Feedback: } Error type: mistranslation Major/minor: major Error location: A meal waited\\
\textbf{Our Error Pinpoint Model: } -5\\
\textbf{BLEURT: }0.703\\
\midrule

\textbf{5th candidate generation: }
A meal took an hour and a half to arrive.\\
\textbf{Feedback: } Error type: no-error Major/minor: no-error Error location: None\\
\textbf{Our Error Pinpoint Model: } 0\\
\textbf{BLEURT: } 0.876\\
\bottomrule
\end{tabular}

\caption{For uphill algorithm, pipeline is stuck at generating "A meal had been waiting for an hour and a half." with BLEURT score $0.738$ for five iterations. For the simulated annealing, despite the model's ability at generating candidates with slightly worse BLEURT score. The pipeline yield global optimum solution at the fifth iteration.}
\label{tab:case_study}
\end{table*}

\begin{table*}[!ht]
\centering

\footnotesize
\begin{tabular}{p{15cm}}
\toprule
\textbf{Prompt for Translation:}
\begin{CJK*}{UTF8}{gbsn}
Translate "一个餐等了一个半小时。" into English. Your translation:
\end{CJK*}\\

\bottomrule
\end{tabular}
\caption{We prompt to obtain initial translation from PALM2.}
\label{tab:mt_examples}
\end{table*}

\begin{table*}[!ht]
\centering

\footnotesize
\begin{tabular}{p{15cm}}
\toprule
\textbf{Prompt for \textsc{IMPROVE}:}
\begin{CJK*}{UTF8}{gbsn}
Translate "一个餐等了一个半小时。" into English. Your translation is "A meal had been waiting for an hour and a half.". Please improve your translation. New translation:
\end{CJK*}\\
\midrule

\textbf{Prompt for \textsc{SCORE-QE}:} 
\begin{CJK*}{UTF8}{gbsn}
Translate "一个餐等了一个半小时。" into English. Your translation is "A meal had been waiting for an hour and a half.". Translation quality is 80 out of 100. Please improve your translation. New translation:
\end{CJK*}\\
\midrule

\textbf{Prompt for \textsc{BINARY-QE}:} 
\begin{CJK*}{UTF8}{gbsn}
Translate "一个餐等了一个半小时。" into English. Your translation is "A meal had been waiting for an hour and a half.". Your translation contains errors. Please improve your translation. New translation:
\end{CJK*}\\
\midrule

\textbf{Prompt for \textsc{ERROR PINPOINT}:} 
\begin{CJK*}{UTF8}{gbsn}
Translate "一个餐等了一个半小时。" into English. Your translation is "A meal had been waiting for an hour and a half.". "A meal had been waiting" is a major mistranslation error. Please improve your translation. New translation:
\end{CJK*}\\
\bottomrule
\end{tabular}

\caption{We include refinement prompts for four different forms of feedback: \textsc{IMPROVE}, \textsc{SCORE-QE}, \textsc{BINARY-QE} and \textsc{ERROR DETECTION}. }
\label{tab:feedback_examples}
\end{table*}

\begin{table*}[!ht]
\centering
\footnotesize
\begin{tabular}{p{15cm}}
\toprule
\textbf{Prompt for error pinpoint model:}\\
\begin{CJK*}{UTF8}{gbsn}
Source translation (Chinese): 一个餐等了一个半小时。 Candidate translation (English): A meal had been waiting for an hour and a half. You are evaluating Chinese-to-English Translation based on source and candidate translations. Your evaluation will contain error type, location and major/minor labels.
\end{CJK*}\\

\textbf{Output for error pinpoint model:}\\
'A meal had been waiting' is a major mistranslation error.\\

\bottomrule
\end{tabular}
\caption{An machine translation example prompt and output we used for our error pinpoint trained from from PALM2. According to our scoring scheme, one major error corresponds to $-5$ weight penalty. The score is $-5$ and we normalize it to $80$ out of $100$.}
\label{tab:instructscore_examples}
\end{table*}

\begin{table*}[!ht]
\centering
\footnotesize
\begin{tabular}{p{15cm}}
\toprule
\textbf{Prompt for error pinpoint model:}\\
"You are evaluating answer based on the passage. Passage: Drag Me to Hell Her boss advises her to demonstrate that she can make tough decisions. An elderly woman, Sylvia Ganush, asks for a third extension on her mortgage payment, and despite Ganushs financial and medical problems, Christine denies her an extension to prove herself. Ganush begs Christine not to repossess her house. Ganush is taken away, accusing Christine of shaming her and swears revenge. In the parking garage Christine is ambushed by Ganush, who is hiding in the back seat. Ganush rips a button from Christines coat and intones words in another language. Later, Christine and her boyfriend Clay meet fortune teller Rham Jas, who tells Christine that she is being haunted by a dark spirit, likely the result of a curse. At home, Christine is attacked by the entity and has nightmares about Ganush. At work the next day, she hallucinates and bleeds profusely from her nose. She leaves the office, and, amid the general panic, Stu steals a file from Christines desk. Christine goes to beg Ganush for forgiveness but discovers that Ganush has died. Christine returns to Jas, who explains that as long as Christine is the owner of an accursed object (the button), she will be haunted by a powerful demon called the Lamia. Drag Me to Hell Drag Me to HellDrag Me to Hell is a 2009 American supernatural horror film co-written and directed by Sam Raimi. The plot, written with his older brother Ivan, focuses on a loan officer, who, because she has to prove to her boss that she can make the hard decisions, chooses not to extend an elderly womans mortgage. In retaliation, the woman places a curse on the loan officer that, after three days of escalating torment, will plunge her into the depths of Hell to burn for eternity. Raimi wrote Drag Me to Hell with his brother, Ivan, before working on the Spider-Man trilogy. The film premiered at the Cannes Film Festival and was released to critical acclaim. It was also a box office success, grossing over \$90 million worldwide. Drag Me to Hell won the award for Best Horror Film at the 2009 Scream Awards and the 2010 Saturn Awards. In 1969, in Pasadena, a couple seeks the aid of the medium Shaun San Dena, saying their son has been hearing evil spirits voices after stealing a silver necklace from a gypsy wagon. San Dena aids the family by carrying out a séance, but they are attacked by an unseen force that drags the boy to Hell. In present-day Los Angeles, bank loan officer Christine is in competition for a promotion with Stu Rubin. Pasadena, California Pasadena, CaliforniaPasadena is a city in Los Angeles County, California, United States, located 10 miles (16 kilometers) northeast of Downtown Los Angeles. The estimated population of Pasadena was 142,647, in 2017, making it the 183rd-largest city in the United States. Pasadena is the ninth-largest city in Los Angeles County. Pasadena was incorporated on June 19, 1886, becoming one of the first cities to be incorporated in what is now Los Angeles County, following the city of Los Angeles (April 4, 1850). It is one of the primary cultural centers of the San Gabriel Valley. The city is known for hosting the annual Rose Bowl football game and Tournament of Roses Parade. In addition, Pasadena is also home to many scientific and cultural institutions, including Caltech, Pasadena City College, Fuller Theological Seminary, ArtCenter College of Design, the Pasadena Playhouse, the Ambassador Auditorium, the Norton Simon Museum, and the USC Pacific Asia Museum. The original inhabitants of Pasadena (a Chippewa word meaning Crown of the Valley) and surrounding areas were members of the Native American Hahamog-na tribe, a branch of the Tongva Nation. They spoke the Tongva language (part of the Uto-Aztecan languages group) and had lived in the Los Angeles Basin for thousands of years. Answer: Drag Me to Hell is a 2009 American supernatural horror film directed by Sam Raimi. The plot focuses on a loan officer who, because she has to prove to her boss that she can make the hard decisions, chooses not to extend an elderly womans mortgage. In retaliation, the woman places a curse on the loan officer that, after three days of escalating torment, will plunge her into the depths of Hell to burn for eternity. The film is set in Pasadena, California and takes place in Pasadenadena, California. Your feedback:"
\\

\textbf{Output for error pinpoint model:}\\
Answer contains a Missing-Major-Auxiliary error, which misses Drag Me to Hell won the award for Best Horror Film at the 2009 Scream Awards and the 2010 Saturn Awards. from passage \\
Answer contains a Missing-Answer error, which misses San Dena aids the family by carrying out a séance, but they are attacked by an unseen force that drags the boy to Hell. from passage \\
directed by Sam Raimi. in the answer is a Irrelevant error. The plot focuses on a loan officer who, because she has to prove to her boss that she can make the hard decisions, chooses not to extend an elderly womans mortgage. in the answer is a Irrelevant error. \\
In retaliation, the woman places a curse on the loan officer that, after three days of escalating torment, will plunge her into the depths of Hell to burn for eternity. in the answer is a Irrelevant error. \\
and takes place in Pasadenadena, California. in the answer is a Redundant error.\\

\bottomrule
\end{tabular}
\caption{A long form QA prompt and output we used for our error pinpoint trained from from PALM2. According to our scoring scheme, one major error corresponds to penalty of $-5$ and one minor error corresponds to penalty of $-1$. The total score is $(-1)*4+(-5)*1=-9$. The normalized score is $64$ out of $100$.}
\label{tab:instructscore_qa_examples}
\end{table*}

\begin{table*}[!ht]
\centering
\footnotesize
\begin{tabular}{p{15cm}}
\toprule
\textbf{Prompt for error pinpoint model:}\\
"You are evaluating a summarization based on question and passage. Passage: It was a cold, dark night…I lay in the corner of the street, my head in a puddle, a smell of what can only be described as death circling my very presence. I had lost count of how many days it had been, of how many faces I’ve seen pass me, of how many feet I’ve watched shuffle aside, trying their hardest to avoid my very existence. Of how many eyes had looked at me, and burnt hatred so deep it became almost intolerable. Sheer disgust, seeping out of the breath of everyone who passed. That is what my life had become. Ever since that one day. When everything I thought I knew was ripped apart in front of me.It was the youngest who took me in. He seemed kind, welcomed me. One of those people that deep down you can tell had a good heart even if they hid it beneath a rough, silent exterior. I lived in his room free to do what I felt, unless the others were around, the older ones. When their voices carried through the door, I was hidden under the bed, its so our friendship will be ours alone I would convince myself, that he wanted me to be all his. But I suppose that should have been the first warning sign. Can a place truly be called your home if you have to hide from its very inhabitants? But sadly, I was naive… no, I chose to ignore it, I was too obsessed, too caught up by this newfound friendship to ever even consider the truth. I thought everything was perfect. Then it happened.  It was late at night, the door slammed behind him as the boy tumbled in. Raised voices instantly burnt through the very walls that surrounded me, through the doors, through everything. I had heard anger in those voices before, the times when I hid, but nothing like this, this was… new. The door crashed open, the boy collapsing to the ground. Eyes bloodshot, his mouth lined with dry, cracked vomit, his shirt, blooded, torn and stained. A sight that I wished I would never see again. As he hit the ground, he looked up at me, but there was no affection any more, just pure emptiness. I heard the voices come closer. There was no time to hide, no time to be hidden. Voices entered the room; eyes were cast down the decrepit shell lying on the floor, then to me. Hatred. Rough hands were placed on me, words crying out that I was to blame, that somehow, this was my entire fault. My protests, my cries of defense, all fell on deaf ears. No matter how much I pleaded, no matter how many times I cried out to them that there was more to me, more I could give to them, it was all to no avail. I was tossed out on the street, my once home fading in my eyesight as I tumbled, seemingly forever, sinking in to my own personal abyss. An icy wind blew threw me, bringing me back from the horrid memory. I rolled onto my side, no longer sure if it was by my own personal doing or if I simply lacked the strength to offer any resistance. This is it, I thought to myself. This is how it is going to end. All the things I could have done, the places I could have seen, and now this will become my final resting place, a blotch in the street, surrounded by the ear wrecking sound of traffic and drunks. Some rest indeed. Through all the noise, I heard footsteps walking towards me. No doubt to impose some form of abuse towards me in my final moments I had thought. But no, a hand rested on my body. Warmth that I had not felt since… the boy? I turned around and stared up, locking eyes with an unfamiliar face. It was irrelevant. This man has picked me up in my time of need. He had saved me. took me to some form of haven for my kind. After I adjusted to the contrast of light, I looked around and saw countless amounts of those in the same position as me. It was amazing. I was no longer a freak. No longer an outcast, I was amongst my own kind at last. The time I spent there was the happiest of my life. On occasions, members of our private little community would be taken away by strangers from the outside, I had lost a few good friends through that, but even then the kind man would always bring in new members, new friends to be made, stories to be heard, connections to be formed. I should have learnt. All good things must come to a pass. The horrendous noise sounded, the noise that sounded the arrival of a stranger. As usual, our host welcomed the man with open arms, allowing him to take his pick of who he wanted. The stranger cast his eyes along the crowd, passing by everyone I knew, his gaze nonchalant to all. All that is, except me. He looked straight at me, and stopped. Eyes bloodshot, his mouth lined with dry, cracked vomit, his shirt, blooded, torn and stained. The memory flooded through me. It was then I realized just how blind I had been. The night that changed it all, it had not been emptiness in the boy’s eyes. It had been lust! And it was lust again the bore its way inside my soul. The stranger had chosen. I wouldn’t be saying goodbye to a friend this time. I would be saying goodbye to everything. The door opened to his building. And a building it would always be, never a home. No place with so much vile hatred seeping out of every crack in the wall, so much murk and gloom pouring out of every shadow, could ever be called a home. He sat in front of me, breathing his horrid breath over my neck, running his fingers down my body, inhaling my breath. I knew straight away, this was never going to end well. He lifted me, I felt weightless and powerless under his grasp. He put his lips around me, and it felt like he drank the very life force from my body. His brown, crusted teeth scraping against me, as his course tongue brought him to the satisfaction he had desired for so long. After what felt like an eternity in the lowest ring of hell, he was done. The door was opened, and for the second time in my life, I was cast out. This time however, there was no remorse. There was nothing. I was resound to my fate.And that’s how I arrived here, lying once again in a gutter. This time, I know there will be no retribution. I can see my life flowing from my body. It’s coming to an end. If only things had been different. If only we really did have medicinal qualities, then maybe, just maybe, this wouldn’t be the end of just one more bottle of Buckfast.  Question: Answer the following question based on the above passage:Summarize what, according to the narrator, led to their being in a gutter. Answer: The narrator was deeply connected to the host, a boy who took the narrator in. After one night out, the narrator came back with dried bloody vomit trailing down his shirt, covered in blood, and the next day the host launched into a tirade of words against the narrator and tore the narrator out of its home. Reduced to a vague memory now, the narrator would recall the chilling last moment they had together. It was a night of deep lust that stained their friendship through the lasting and eternal pains of betrayal and lust. Your Feedback:"\\
\textbf{Output for error pinpoint model:}\\
This response contains a major coherence error. The sentence is written in a confusing way. The narrator realizes the lust in the boy's eyes, and their last night is horrendous for it. Then, the narrator is cast out again.\\

\bottomrule
\end{tabular}
\caption{A topical summarization example prompt and output we used for our error pinpoint trained from from PALM2, where a summarization example is based on a particular question. According to our scoring scheme, one major error corresponds to $2$, at $1$ to $5$ scale. The normalized score is $40$ out of $100$.}
\label{tab:instructscore_summ_examples}
\end{table*}

\section{Additional Results}
\label{appendix:additional_results}
Table~\ref{tab:charf1_all} contains the character-level evaluation of the span tagging models, AutoMQM and our error detection feedback model, on the full WMT'22 dataset.
XCOMET is omitted because the non-news sections of the dataset were used for training and is thus not a fair comparison.

\begin{table}[t]
    \centering
    \begin{adjustbox}{width=\columnwidth}
    \begin{tabular}{lcccccc}
        \toprule
        \multirow{2}{*}{\bf Metric} & \multicolumn{3}{c}{\bf en-de} & \multicolumn{3}{c}{\bf zh-en} \\
        \cmidrule{2-7}
        & \bf P & \bf R & \bf F$_1$ & \bf P & \bf R & \bf F$_1$ \\
        \midrule
        AutoMQM (Bison) & 0.06 & 0.64 & 0.11 & 0.14 & 0.21 & 0.17 \\
        Error Pinpoint & 0.29 & 0.20 & 0.24 & 0.29 & 0.36 & 0.32 \\
        \bottomrule
    \end{tabular}
    \end{adjustbox}
    \caption{Character-level precision/recall/F$_1$ of different reference-free error span tagging models on the full WMT'22 dataset.}
    \label{tab:charf1_all}
\end{table}

\begin{table}[t]
    \centering
    \begin{adjustbox}{width=0.90\columnwidth}
    \begin{tabular}{lcc}
        \toprule
        \multirow{1}{*}{\bf Mistral MoE} & \multicolumn{1}{c}{\bf WMT22 Zh-En} & \multicolumn{1}{c}{\bf WMT22 En-De} \\
        \midrule
        0-shot & 0.778 & 0.779 \\
        Improve & 0.781 & 0.801 \\
        Score & 0.775 & 0.800 \\
        Binary & 0.778 & 0.804 \\
        Fine-grained & 0.786 & 0.812 \\
        \bottomrule
    \end{tabular}
    \end{adjustbox}
    \caption{\method's performance on Mistral MoE \cite{jiang2024mixtral} compared to coarse feedback, measured by metricX.}
    \label{tab:mistral_moe_metricx}
\end{table}

\begin{table}[t]
    \centering
    \begin{adjustbox}{width=0.90\columnwidth}
    \begin{tabular}{lcc}
        \toprule
        \multirow{1}{*}{\bf PALM2} & \multicolumn{1}{c}{\bf WMT22 Zh-En} & \multicolumn{1}{c}{\bf WMT22 En-De} \\
        \midrule
        0-shot & 0.747 & 0.818 \\
        Improve & 0.757 & 0.797 \\
        Score & 0.757 & 0.813 \\
        Binary & 0.757 & 0.813 \\
        Fine-grained & 0.759 & 0.823 \\
        \bottomrule
    \end{tabular}
    \end{adjustbox}
    \caption{\method's performance on PALM2 compared to coarse feedback, measured by COMET22.}
    \label{tab:palm2_comet22}
\end{table}

\begin{table}[!htp]
    \centering
    \begin{adjustbox}{width=\columnwidth}
    \begin{tabular}{lcccccc}
        \toprule
        \multirow{2}{*}{\bf Search Algorithm} & \multicolumn{2}{c}{\bf Zh-En} & \multicolumn{2}{c}{\bf En-De} & \multicolumn{1}{c}{\bf ASQA} & \multicolumn{1}{c}{\bf Summ}\\
        \cmidrule{2-7}
        & \bf 22 & \bf 23 & \bf 22 & \bf 23 & \bf QA & \bf Sum\\
        \midrule
        \textsc{Zero-shot} & 67.6 & 67.3 & 79.0 & 77.0 & 18.3 & 26.1 \\
        \midrule
        \textsc{Always Accept (1)} & 69.3 & 68.4 & 79.4 & 77.5 & 26.2 & 27.0\\
        \textsc{Greedy Uphill (1)} & 69.1 & 68.8 & 79.6 & 77.1 & 25.5 & 27.5\\ 
        \textsc{Sim. Annealing (1)} & 69.2 & 68.4 & 79.7 & 77.5 & 25.4 & 27.5\\  %
        \midrule
        \textsc{Always Accept (5)} & 69.9 & 68.3 & 80.0 & 78.1 & 25.4 & 27.4\\  
        \textsc{Greedy Uphill (5)} & 69.6 & 68.9 & 80.1 & 77.3 & 25.7 & 27.4\\  
        \textsc{Sim. Annealing (5)} & \bf 70.1 & \bf 69.2 & \bf 81.0 & \bf 78.4 & \bf 26.4 & \bf 28.3\\ 
        \bottomrule
    \end{tabular}
    \end{adjustbox}
    \caption{We include iterative refinement results from three search algorithms: 1) Always Accept 2) Greedy Uphill 3) Simulated Annealing for $1$ iteration and $5$ iterations. Different from Table \ref{tab:main_results}, all search algorithms are performed with top-k sampling at each step and we report results on examples that are marked as "errors" by our feedback model.}
    \label{tab:search_results}
\end{table}
\label{tab:ier_refine_table}


\end{document}